\documentclass[10pt,twocolumn,letterpaper]{article}

\usepackage[pagenumbers]{cvpr} 

\definecolor{cvprblue}{rgb}{0.21,0.49,0.74}
\usepackage[pagebackref,breaklinks,colorlinks,allcolors=cvprblue]{hyperref}

\usepackage{multirow}
\usepackage{subcaption}
\usepackage{booktabs}
\usepackage{pifont}

\usepackage{xcolor}
\definecolor{MethodBlue}{HTML}{0072B2}
\definecolor{MethodOrange}{HTML}{E69F00}
\definecolor{MethodTeal}{HTML}{009E73}

\usepackage{tikz}
\newcommand{\cdotcolor}[2][0.7ex]{%
  \tikz[baseline=-0.6ex]\draw[fill=#2,draw=none] (0,0) circle (#1);%
}
\newcommand{\BlueDot}{\cdotcolor{MethodBlue}}
\newcommand{\OrangeDot}{\cdotcolor{MethodOrange}}
\newcommand{\TealDot}{\cdotcolor{MethodTeal}}

\newcommand{\circnum}[1]{%
  \tikz[baseline=(c.base)]\node (c)
    [draw, circle, inner sep=0.25ex, line width=0.4pt]{\scriptsize #1};}

\usepackage[final]{microtype} 
\microtypesetup{
    expansion=true,
    stretch=10,
    shrink=30     
}

\renewcommand{\paragraph}[1]{\vspace{.5em}\noindent\textbf{#1}}

\usepackage[table]{xcolor}
\definecolor{OursRow}{RGB}{222,235,247}
\usepackage{array}
\newcolumntype{G}{>{\color{gray}\arraybackslash}c} 

\usepackage{bbm}
\def\paperID{****} 
\def\confName{CVPR}
\def\confYear{2026}

\title{Unified Primitive Proxies for Structured Shape Completion}

\author{ Zhaiyu Chen$^{1,2}$ \quad\quad Yuqing Wang$^1$ \quad\quad Xiao Xiang Zhu$^{1,2}$\\[0.20cm]
  {
    $^{1}$Technical University of Munich \quad\quad
    $^{2}$Munich Center for Machine Learning
  }
}

\begin{document}
\twocolumn[{
\maketitle
\begin{center}
\vspace{-2.0em}
\includegraphics[width=0.99\linewidth]{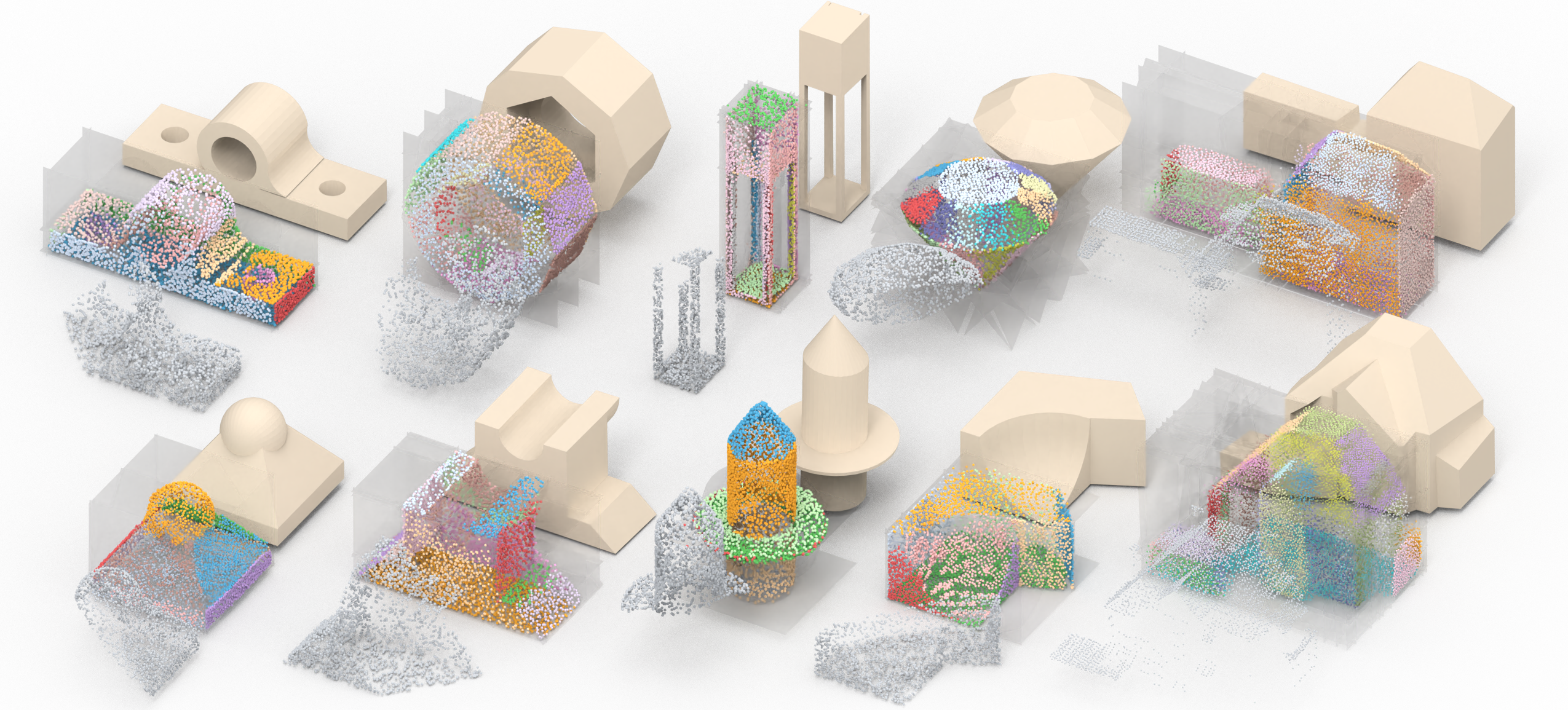}
\end{center}
\vspace{-0.6cm}
\captionsetup{type=figure}
\captionof{figure}{%
\textbf{
}
We present \textit{UniCo}, a structured shape completion model that, given a partial scan, jointly predicts a complete set of quadratic primitives with geometry, semantics, and inlier membership. The predicted primitives are assembly-ready for surface reconstruction.
}\label{fig:teaser}
\vspace{0.4cm}
}]

\begin{abstract}
Structured shape completion recovers missing geometry as primitives rather than as unstructured points, which enables primitive-based surface reconstruction. Instead of following the prevailing cascade, we rethink how primitives and points should interact, and find it more effective to decode primitives in a dedicated pathway that attends to shared shape features. Following this principle, we present UniCo, which in a single feed-forward pass predicts a set of primitives with complete geometry, semantics, and inlier membership. To drive this unified representation, we introduce primitive proxies, learnable queries that are contextualized to produce assembly-ready outputs. To ensure consistent optimization, our training strategy couples primitives and points with online target updates. Across synthetic and real-world benchmarks with four independent assembly solvers, UniCo consistently outperforms recent baselines, lowering Chamfer distance by up to 50\% and improving normal consistency by up to 7\%. These results establish an attractive recipe for structured 3D understanding from incomplete data. Project page: \url{https://unico-completion.github.io}.
\end{abstract}
    
\section{Introduction}
\label{sec:intro}

Occlusions and limited sensor coverage often leave 3D scans incomplete. Completing the missing geometry allows robots to plan stable grasps, enables autonomous vehicles to perceive hidden traffic, and supports digitizing heritage artifacts without repeated acquisitions~\cite{tabib2023defi,tsesmelis2024re,zheng2024towards,varley2017shape}. 

Despite advances, most shape completion methods still optimize pointwise discrepancies~\cite{yuan2018pcn,yu2021pointr,yu2023adapointr} or their variants~\cite{wu2021density,lin2023hyperbolic,lin2023infocd}. These objectives capture local geometry but convey little about the structural regularities required by many downstream tasks~\cite{botsch2010polygon, koch2019abc}. In contrast, primitive assembly models the surface as a compact, topologically consistent collection of parametric primitives for structured, interpretable geometry~\cite{nan2017polyfit,bauchet2020kinetic,sulzer2024concise,jiang2023primfit}. However, the common recipe of completing first and assembling later is underconstrained, because assembly solvers expect structured input that pointwise completion does not provide. It is therefore preferable to predict structure jointly with completion.

Toward structured shape completion that directly supports primitive assembly, a straightforward approach is a two-stage cascade that first regresses primitive parameters and then enforces inlier points, typically limited to plane-only primitives~\cite{chen2025parametric}. In practice, this rigid formulation tends to overfit well-supported regions and degrade when evidence is sparse. It can also propagate early errors in primitive count or parameters into the association step, which weakens later supervision. These limitations motivate a formulation in which point completion and primitive inference are optimized in a more coordinated way.

\textit{How can primitives be optimized more effectively?} We follow three design principles: \circnum{1} \textit{Coordinated pathways.} Point completion and primitive inference are driven by different supervision signals, since the former benefits from pointwise guidance whereas the latter relies on discrete and relational cues. We therefore let completion run in its own pathway and decode primitives in parallel from shared features. \circnum{2} \textit{Unified representation.} Structural information is dispersed across the shared features, which makes coordination nontrivial. We introduce primitive proxies that attend to these shared features and provide a unified representation that binds evidence to candidate primitives. \circnum{3} \textit{Consistent optimization.} Early in training, the predicted point distribution is not accurate enough to support reliable primitive membership. We therefore update primitive targets online, while keeping the matching permutation-invariant, so that both pathways maintain stable training dynamics.

Following these principles, we present UniCo, a structured shape completion model that predicts assembly-ready quadratic primitives with complete geometry, semantics, and inlier membership in a single pass. UniCo is purpose-built to enable reliable primitive assembly from challenging incomplete data. To summarize, our contributions are:
\begin{itemize}
    \item \textit{Formulation.} We rethink how primitives and points should be coordinated and introduce UniCo, a structured shape completion model that jointly optimizes both, producing assembly-ready primitives in one pass.
    \item \textit{Representation.} We present primitive proxies as learnable queries over shared features that produce a unified primitive representation and jointly drive geometry, semantics, and inlier membership predictions.
    \item \textit{Optimization.} We develop a training strategy with online target updates for consistent support between primitives and points as predictions evolve.
\end{itemize}

\noindent These contributions realize our design principles in a single versatile solution. UniCo supports multiple primitive families and includes a planar variant for cases dominated by planar structures. On three benchmarks covering synthetic and real scans, and evaluated with four assembly solvers, UniCo consistently outperforms recent baselines, lowering Chamfer distance by up to 50\% and improving normal consistency by up to 7\%. We hope these findings stimulate further work on structured 3D understanding from incomplete data.
\section{Related Work}
\label{sec:relatedwork}

\paragraph{3D shape completion.}
Early approaches to shape completion used volumetric CNNs, but voxel representations suffer from discretization artifacts and high memory costs at fine resolutions~\cite{wu2015, dai2017, xie2020grnet, han2017high}. The introduction of PointNet enabled direct processing of unordered point sets~\cite{qi2017pointnet}, which led to a broad family of point-based completion networks~\cite{qi2017pointnet++, wang2019dynamic, thomas2019kpconv}. Most recent methods, including PoinTr~\cite{yu2021pointr}, AdaPoinTr~\cite{yu2023adapointr}, ODGNet~\cite{cai2024orthogonal}, SymmComplete~\cite{yan2025symmcompletion}, and others~\cite{yuan2018pcn, yang2018foldingnet, xiang2021snowflakenet, zhao2022seedformer, tang2022lake, yan2022fbnet, chen2023anchorformer, zhu2023csdn, tesema2023point, lee2024proxyformer}, still minimize pointwise discrepancies or their variants~\cite{wu2021density, lin2023hyperbolic, lin2023infocd}, so they recover local geometry but leave higher-level structure underconstrained. PaCo moves toward structured shape completion by first predicting plane parameters and then enforcing inlier membership in a cascade, yet this design remains sensitive to sparse evidence and early errors, and is restricted to plane-only primitives~\cite{chen2025parametric}. In contrast, our approach unifies shape completion and structural reasoning in a single network with coordinated pathways, enabling consistent optimization across multiple primitive families.

\paragraph{Primitive assembly.}
Unlike generic surface reconstruction that targets dense meshes~\cite{kazhdan2013screened, erler2020points2surf, huang2022neural, huang2024surface}, primitive assembly reconstructs surfaces by arranging primitives under geometric and topological constraints. Although recent neural approaches explore learned decompositions and constructive modeling~\cite{tulsiani2017learning, chen2020bspnet, ren2021csg, yu2022capri, li2023secad, liu2024point2cad}, primitive assembly remains the practical standard when topology control, editability, and compatibility are required~\cite{botsch2010polygon, kaiser2019survey, chen2024polygnn, chen2022points2poly, chen2025abspy}. Representative solvers include PolyFit~\cite{nan2017polyfit}, KSR~\cite{bauchet2020kinetic}, and COMPOD~\cite{sulzer2024concise} for polygonal surfaces, and PrimFit~\cite{jiang2023primfit} for more general primitive families. However, their performance degrades under partial observations because assembly depends on reliable, complete primitives, which motivates learning structured representations from partial scans that supply assembly-ready primitives.

\paragraph{Primitive extraction.}
Primitive extraction is often cast as instance segmentation. Traditional pipelines rely on geometric model fitting and constraints, but lack learned inductive biases~\cite{schnabel2007efficient, li2011globfit, yu2022finding}. Learning-based methods later adopted a clustering paradigm, where a shared backbone first produces pointwise features and a subsequent stage groups points into primitive instances~\cite{li2019supervised, sharma2020parsenet, le2021cpfn, yan2021hpnet}. This paradigm led benchmarks for several years~\cite{jiang2020pointgroup, lahoud20193d, chen2021hierarchical, liang2021instance, vu2022softgroup}. More recent Transformer-based models predict masks directly with instance queries and avoid hand-crafted grouping~\cite{schult2022mask3d, sun2023superpoint, lai2023mask, lu2023query}. However, they rarely encode primitive-specific priors and typically assume complete, fixed point sets, which hinders transfer to completion settings with inconsistent targets. In contrast, we infer primitives in tandem with completion, jointly predicting memberships from incomplete scans while reasoning about missing geometry.
\begin{figure*}[ht]
  \centering
  \includegraphics[width=0.99\linewidth]{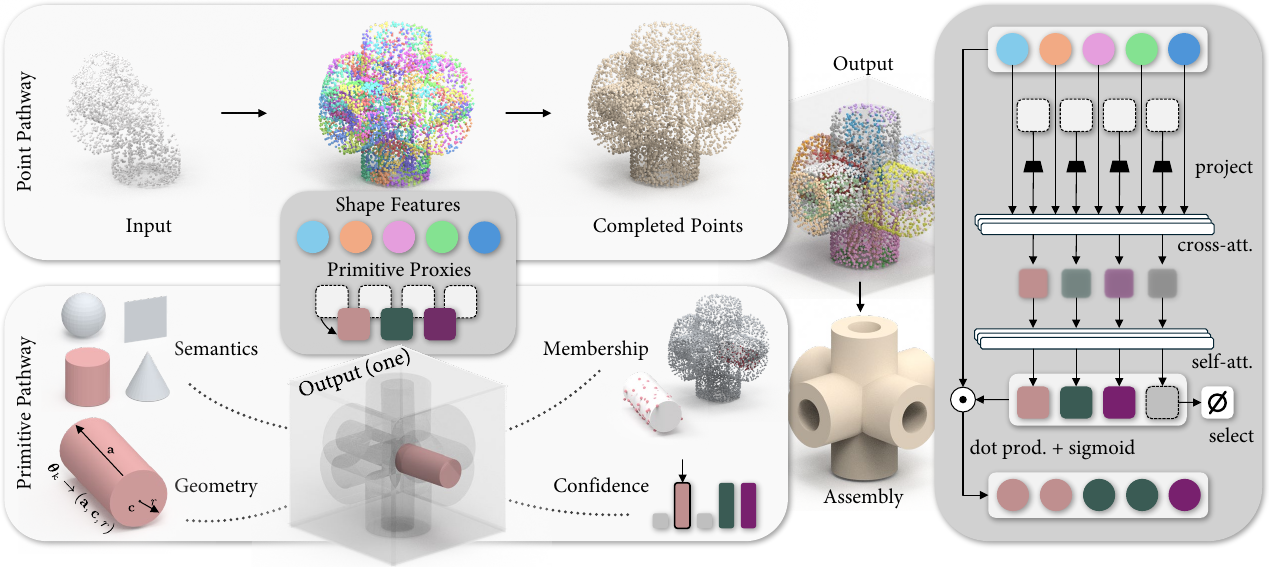}
  \caption{\textbf{Architecture of UniCo.} Shape features from a partial point cloud feed two coordinated pathways. The \emph{point pathway} decodes dense completed points. The \emph{primitive pathway} uses primitive proxies that attend to the shared features and predict primitive semantics, geometry, inlier membership, and a confidence score used at inference to select valid primitives. The selected primitives are assembly-ready.}
  \label{fig:architecture}
\vspace{-1em}
\end{figure*}

\section{Method}
\label{sec:method}

Given a partial point cloud, we complete the shape by jointly predicting a point set and a primitive set with geometry, semantics, and inlier membership, yielding an assembly-ready representation of the complete object.
\cref{fig:architecture} illustrates the architecture. We extract shape features from the input points. The point pathway decodes dense points, while the primitive pathway contextualizes a fixed set of primitive proxies from the shared features to predict candidate primitives. During training, we update primitive targets online and use permutation-invariant matching to keep supervision aligned with evolving predictions. At inference, confidence scores select the valid subset of primitives.

\subsection{Coordinated Pathways}

The point pathway reconstructs dense geometry, while the primitive pathway infers structural elements that abstract the object into parametric primitives. Both pathways share the same latent features $\mathcal{T} = \{\mathbf{t}^u\}_{u=1}^U$, which promotes consistency between fine-grained point completion and higher-level primitive prediction.

\paragraph{Point pathway.}
We instantiate the point pathway \(f_{\mathrm{point}}\) with AdaPoinTr~\cite{yu2023adapointr}, though in principle any pointwise shape completion network could be used. This pathway recovers a local point patch from each feature \(\mathbf{t}^u\) and aggregates them into the completed points \(\mathcal{\hat Y} \subset \mathbb{R}^3\):
\begin{equation}
\hat{\mathcal{Y}} 
= f_{\mathrm{point}}(\mathcal{T})
= \bigcup_{u=1}^{U} \hat{\mathcal{Y}}^{u}, \quad \hat{\mathcal{Y}}^{u} = \{\hat{\mathbf{y}}_{j}^u\}_{j=1}^{J}, 
\end{equation}
\noindent where \(\hat{\mathbf{y}}_{j}^u\) denotes a point in the patch decoded from \(\mathbf{t}^u\).

\paragraph{Primitive pathway.}
In parallel, the primitive pathway \(f_{\mathrm{primitive}}\) employs a set of primitive proxies \(\mathcal{R} \subset \mathbb{R}^d\), which are contextualized against the same features:
\begin{equation}
{\mathcal{R}} = f_{\mathrm{primitive}}(\mathcal{T}, \mathcal{R}^{(0)}) 
= \{{\mathbf{r}}_{k}\}_{k=1}^{K},
\label{eq:primitive_pathway}
\end{equation}
where \(\mathcal{R}^{(0)}\) denotes the initialized proxies and \({\mathcal{R}}\) the contextualized ones. The embeddings are subsequently processed by dedicated prediction heads.

\subsection{Primitive Proxies}
\label{sec:proxies}
We introduce primitive proxies, learnable queries that aggregate dispersed structural cues from the shape features into unified primitive-level representations. A fixed set of proxies is contextualized with the shared shape features and then decoded by the geometry, semantics, and membership heads.

\paragraph{Contextualization.}
The primitive proxies are initialized as queries \(\mathcal{R}^{(0)}\). At layer \(l\), they attend to the shared shape features \(\mathcal{T}\) and then interact among themselves:
\begin{equation}
  \mathcal{R}^{(l)} = \operatorname{self-att}\big(\operatorname{cross-att}\big(\mathcal{R}^{(l-1)}, \operatorname{MLP}(\mathcal{T})\big)\big).
\end{equation}
The final contextualized proxies \({\mathcal{R}}=\{{\mathbf{r}}_{k}\}_{{k}=1}^K\)\ are shared across the prediction heads. \cref{fig:architecture} illustrates this process.

\paragraph{Semantics.} 
For mixed-type settings, an MLP classification head predicts the type of each primitive candidate:
\begin{equation}
\label{eq:type-softmax}
\boldsymbol{\pi}_{k} = \operatorname{softmax}\big(\operatorname{MLP}({\mathbf{r}}_{k})\big),
\end{equation}
where \(\boldsymbol{\pi}_{k}\) is a categorical distribution over five classes (plane, cylinder, sphere, cone, \(\emptyset\)), with \(\emptyset\) denoting noncontributing candidates. The formulation can extend to additional primitive families, provided the downstream solver supports them. In plane-only settings, \cref{eq:type-softmax} reduces to a binary classifier.

\paragraph{Membership.} 
Each primitive candidate predicts its inlier subset among the completed points. Given a primitive proxy \({\mathbf{r}}_{k}\) and a shape feature \(\mathbf{t}^u\), we compute their pairwise similarity in a shared latent space:
\begin{equation}
\label{eq:inlier-score}
{m}_{k}^u
= \operatorname{sigmoid}\big(\langle \operatorname{MLP}({\mathbf{r}}_{k}), \operatorname{MLP}(\mathbf{t}^u) \rangle\big),
\end{equation}
where both \(\operatorname{MLP}(\cdot)\) project into the same latent space, and \(\langle \cdot , \cdot \rangle\) denotes the dot product. Inlier points for primitive \(k\) are then obtained by thresholding:
\begin{equation}
\begin{aligned}
\hat{\mathcal{I}}_{k} &= \{\, u\mid {m}_{k}^u > 0.5 \,\}.
\end{aligned}
\end{equation}
All similarity scores are stacked into a membership matrix \({\mathbf{M}} \in [0,1]^{K \times U}\) for optimization.

\paragraph{Geometry.} 
The geometry head maps each contextualized proxy \({\mathbf{r}}_{k}\) to quadric parameters, providing a unified homogeneous parametrization for common primitives:
\(
\boldsymbol{\theta}_{k} := \mathbf{A}_{k} = \operatorname{MLP}({\mathbf{r}}_{k}).
\)
The corresponding surface is defined as: 
\begin{equation}
\label{eq:quadric}
\begin{aligned}
&\mathbf{x}^\top \mathbf{A}_{k} \mathbf{x} = 0, \quad 
\mathbf{A}_{k} = \mathbf{A}_{k}^\top \in \mathbb{R}^{4\times 4},
\end{aligned}
\end{equation}
where \(\mathbf{x}\) denotes the homogeneous coordinates of a point on the surface.
For plane-only settings, we set \([\mathbf{A}_k]_{1:3,\,1:3}= \mathbf{0}\) to avoid ambiguity. We also derive the dense point geometry for each primitive using its predicted membership:
\begin{equation}
\hat{\mathcal{Y}}_{k} = \bigcup_{u \in \hat{\mathcal{I}}_{k}} \hat{\mathcal{Y}}^{u}.
\end{equation}

\subsection{Optimization}
\label{sec:optimization}

Unlike standard primitive learning on fixed point sets with direct membership supervision, our predicted points evolve during training, so a fixed point-to-membership correspondence would be ill-defined. We therefore induce membership supervision online: at each iteration we assign ground truth primitive labels to the current predictions and train on these induced labels. To align the unordered set of predicted primitives with targets, we compute the loss after permutation-invariant matching. As training progresses, both the induced memberships and the matching are updated, resulting in a self-consistent optimization loop.

\paragraph{Online targets.}
We first assign labels to the predicted points. Given the ground-truth point set 
\(\mathcal{Y}=\{\mathbf{y}_i\}_{i=1}^{N}\) with primitive labels \(\mathcal{P}=\{p_i\}_{i=1}^{N}\), where \(p_i \in \{1,\dots,G\}\) denotes the primitive index of point \(i\). Each predicted point \(\hat{\mathbf{y}}_{j}^u\) takes the label of its nearest ground-truth neighbor:
\begin{equation}
\hat{p}_{j}^u
=
p_{i^{*}}, \quad
{i^{*}}=
{\arg\min_{i}\, \lVert \hat{\mathbf{y}}_{j}^u - \mathbf{y}_i \rVert_2 }.
\end{equation}
The patch-level primitive label is then obtained by a majority vote of these assigned point labels:
\begin{equation}
\hat{\mathcal{P}}^{u}
=
\arg\max_{g}
\sum_{j=1}^{J} 
\mathbbm{1}\{\hat{p}_{j}^{u} = g\},
\end{equation}
where \(\mathbbm{1}\{\cdot\}\) is the indicator function and \(g\) indexes a primitive. 
For each primitive, we collect the patches assigned to it:
\begin{equation}
{\mathcal{I}}_{g} = \{\, u \mid \hat{\mathcal{P}}^{u} = g \,\},
\end{equation}
and use these sets as online targets to supervise the primitive-specific predictions. The targets are recomputed at every iteration, allowing assignments and network parameters to be optimized jointly during training.

\paragraph{Matching and losses.} 
To align the unordered predicted primitives with ground-truth primitives, we establish a pairwise cost and solve a bipartite assignment:
{\small
\begin{equation}
\label{eq:matching}
\begin{aligned}
\operatorname{cost}(k,g)
&=\;-\alpha_{\mathrm{1}}
\underbrace{
  \log \boldsymbol{\pi}_{k}[c_g]
}_{\text{semantics}}
\;+\;
\alpha_{\mathrm{2}}\,
\underbrace{
  \bigl(\operatorname{CE} + \operatorname{Dice}\bigr)\bigl({\mathbf{M}}_{k}, {\mathcal{I}}_g\bigr)
}_{\text{membership}}
\\[0.6ex]
&\quad+\;
\alpha_{3}\,
\underbrace{  \bigl[\operatorname{CD}\!\big(\hat{\mathcal{Y}}_{k},\,\mathcal{Y}_{g}\big)
    \;+\;
    \lambda
    \bigl\lVert \boldsymbol{\theta}_k - \boldsymbol{\theta}_{g} \bigr\rVert_1
  \bigr]
}_{\text{geometry}}.
\end{aligned}
\end{equation}}

\noindent Here, \(\alpha_1, \alpha_2, \alpha_3\) and \(\lambda\) are balancing weights. The semantic term encourages correct primitive type prediction. The membership term enforces point-primitive membership consistency using cross-entropy and Dice losses~\cite{deng2018learning}. The geometry term aligns predicted inliers with the ground truth in both Chamfer distance and parameters. The optimal bipartite matching \(\mathcal{M}\) is obtained with the Hungarian algorithm~\cite{kuhn1955hungarian}.
The overall loss combines the matched primitive costs with an object-level distance from the point pathway:
\begin{equation}
\label{eq:total}
\mathcal{L}_{\mathrm{total}}
\;=\;
\sum_{(k^*,g^*)\in \mathcal{M}} \mathrm{cost}(k^*,g^*)
\;+\;
\operatorname{CD}\!\big(
\hat{\mathcal{Y}},\,
\,\mathcal{Y}
\big). 
\end{equation}
Unmatched predictions are downweighted via the semantic term to mitigate class imbalance.

\paragraph{Inference.}
At inference time, inspired by practices in instance segmentation~\cite{cheng2021per, schult2022mask3d}, we score each predicted primitive by combining its semantic confidence with the reliability of its inliers:

\begin{equation}
\label{eq:final-score}
\begin{aligned}
&s_{k}
=
\boldsymbol{\pi}_{k}[\hat{c}_{k}]
\cdot
\frac{1}{|\hat{\mathcal{I}}_{k}|}
\sum_{u\in \hat{\mathcal{I}}_{k}} {m}_{k}^u, \\[0.5ex]
&\text{where} \quad
\hat{c}_{k}
=
\arg\max_{c\neq\emptyset}\boldsymbol{\pi}_{k}[c].
\end{aligned}
\end{equation}
Primitives with \(s_{k} > 0.5\) are retained and passed to the downstream assembly solver. Architectural and implementation details are provided in the \underline{Appendix}.
\begin{figure*}[t]
  \centering
  \includegraphics[width=0.99\linewidth]{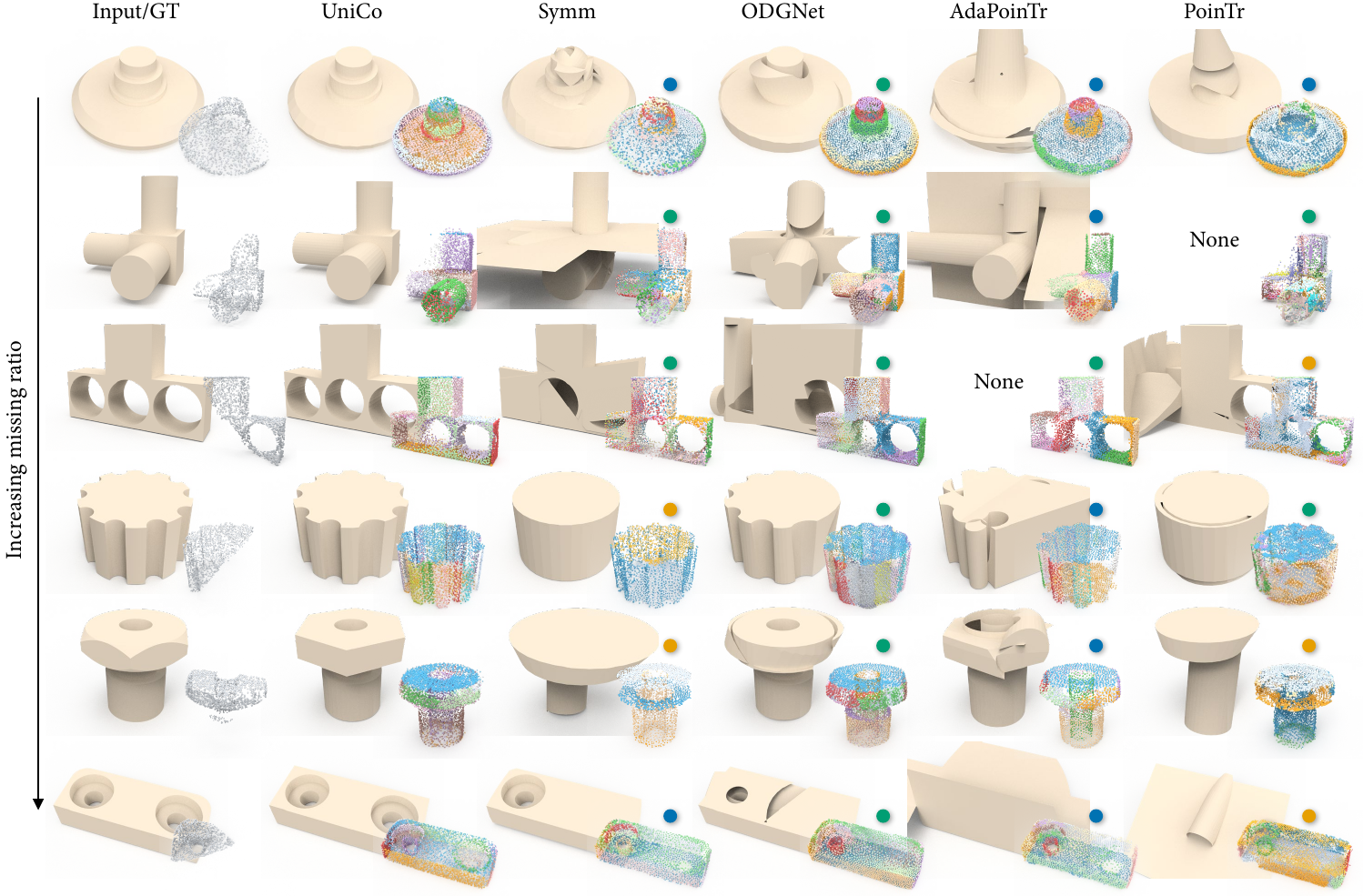}
  \vspace{-0.3em}
  \caption{\textbf{Comparison to completion baselines on \textit{ABC-multi}.} For each baseline, completed points are paired with its best-performing primitive extractor (\BlueDot~RANSAC~\cite{schnabel2007efficient}, \TealDot~HPNet~\cite{yan2021hpnet}, \OrangeDot~PTv3~\cite{wu2024ptv3}). UniCo recovers extractor-free, assembly-ready primitive structures.}
  \label{fig:completion}
\vspace{-0.5em}
\end{figure*}

\section{Experiments}
\label{sec:experiments}

\subsection{Setup and Protocol}

\paragraph{Datasets.}
We evaluate on three datasets for a comprehensive assessment. \textit{ABC-multi} is a curated subset of 30{,}000 watertight CAD models from the ABC dataset~\cite{koch2019abc}, with 5{,}000 reserved for evaluation. It spans planes, cylinders, spheres, and cones, stressing completion with mixed primitive types. \textit{ABC-plane}~\cite{chen2025parametric} is an existing plane-only dataset with over 15{,}000 CAD models. We use it to assess behavior in plane-dominant settings and to compare with polygonal pipelines. For real-world evaluation, we use \textit{Building-PCC}~\cite{gao2024building}, an airborne LiDAR dataset of about 50{,}000 urban buildings with realistic noise and occlusions. Following prior works~\cite{yu2023adapointr, chen2025parametric}, ABC-multi and ABC-plane use 2{,}048 input points down-sampled from partial points at 25\%, 50\%, and 75\% incompleteness, and 8{,}192 target points per shape, while Building-PCC uses native point counts.

\paragraph{Baselines and solvers.}
We compare three groups of baselines. For \textit{completion}, we evaluate GRNet~\cite{xie2020grnet}, PoinTr~\cite{yu2021pointr}, AdaPoinTr~\cite{yu2023adapointr}, ODGNet~\cite{cai2024orthogonal}, SymmComplete~\cite{yan2025symmcompletion}, and PaCo~\cite{chen2025parametric}. Unless stated otherwise, all methods are trained to convergence. For \textit{primitive extraction}, on ABC-multi we fit primitives to completed points using RANSAC~\cite{schnabel2007efficient}, HPNet~\cite{yan2021hpnet}, and PTv3~\cite{wu2024ptv3}, followed by PrimFit~\cite{jiang2023primfit} for assembly. On ABC-plane and Building-PCC, we use GoCoPP~\cite{yu2022finding} to extract primitives and then apply PolyFit~\cite{nan2017polyfit}, KSR~\cite{bauchet2020kinetic}, and COMPOD~\cite{sulzer2024concise} for assembly. For \textit{reconstruction}, we include BSP-Net~\cite{chen2020bspnet} and Point2CAD~\cite{liu2024point2cad}, evaluated on both raw and completed points. For multi-stage pipelines (\eg, completion followed by reconstruction), we compose best models at each stage to obtain competitive baselines.

\paragraph{Metrics.}
Unless stated otherwise, evaluations are conducted on reconstructed meshes using Chamfer distance (CD), Hausdorff distance (HD), and normal consistency (NC). We also report the solver failure rate (FR), defined as the fraction of samples for which reconstruction fails. Additional metrics and setup details are provided in the \underline{Appendix}.

\subsection{Mixed-Type Primitive Results}

\begin{table*}[htbp]
\centering
\caption{\textbf{Comparison with completion baselines on \textit{ABC-multi}.} Best scores are \textbf{bold}, second best are \underline{underlined}. Baseline methods require a separate primitive extractor, whereas UniCo produces structured completion directly. All results are reconstructed with the same PrimFit solver~\cite{jiang2023primfit}. CD, HD, and FR values are scaled by 100.}
\vspace{-0.3em}
\footnotesize
\resizebox{0.99\textwidth}{!}{
\begin{tabular}{l
                GG!{\vrule width .6pt}
                cccc!{\vrule width .6pt}
                cccc!{\vrule width .6pt}
                cccc}
\toprule
\multirow{2}{*}{\raisebox{-0.5ex}{Method}} &
\multicolumn{2}{c}{\color{gray}Pointwise} &
\multicolumn{4}{c}{RANSAC \cite{schnabel2007efficient}} &
\multicolumn{4}{c}{HPNet \cite{yan2021hpnet}}  & \multicolumn{4}{c}{PTv3 \cite{wu2024ptv3}}
\\
\cmidrule(lr){2-3}\cmidrule(lr){4-7}\cmidrule(lr){8-11}\cmidrule(lr){12-15}
& CD & F1
& CD $\downarrow$ & HD $\downarrow$ & NC $\uparrow$ & FR $\downarrow$
& CD $\downarrow$ & HD $\downarrow$ & NC $\uparrow$ & FR $\downarrow$
& CD $\downarrow$ & HD $\downarrow$ & NC $\uparrow$ & FR $\downarrow$ \\
\midrule

GRNet \cite{xie2020grnet}
& 1.177 & 0.565
& 18.61 & 24.71 & 0.543 & 100.00
& 18.61 & 24.71 & 0.543 & 100.00
& 18.61 & 24.71 & 0.543 & 100.00\\

PoinTr \cite{yu2021pointr}
&0.719 & 0.764
& 6.58 & 22.62 & 0.814 & 17.07
& 7.56 & 21.38 & 0.779 & 26.70
& 6.58 & 19.41 & 0.791 & 11.27\\

AdaPoinTr \cite{yu2023adapointr}
& \underline{0.625} & 0.824
& 6.81 & \underline{21.24} & 0.815 & 19.77
& 4.41 & \underline{13.36} & 0.872 & 8.97 
& 5.58 & 16.80 & 0.821 & 9.83\\

ODGNet \cite{cai2024orthogonal}
& 0.632 & \textbf{0.850}
& \underline{4.80} & 22.15 & \underline{0.868} & \textbf{0.39}
& \underline{4.33} & 13.63 & \underline{0.873} & \underline{7.41}
& \underline{5.48} & \underline{16.79} & \underline{0.823} & \underline{9.18}\\

SymmComplete \cite{yan2025symmcompletion}
& \textbf{0.487} & \underline{0.825}
& 7.54 & 21.42 & 0.796 & 26.13
& 4.57 & 13.58 & 0.865 & 9.84 
& 5.93 & 17.58 & 0.812 & 11.50\\

\rowcolor{OursRow}
UniCo (ours)
& 0.686 & 0.799
& \textbf{2.18} & \textbf{7.53} & \textbf{0.935} & \underline{1.49}
& \textbf{2.18} & \textbf{7.53} & \textbf{0.935} & \textbf{1.49} 
& \textbf{2.18} & \textbf{7.53} & \textbf{0.935} & \textbf{1.49}\\
\bottomrule
\end{tabular}}
\label{tab:abc_multi}
 \vspace{-1em}
\end{table*}

\paragraph{Structured \vs pointwise.}
On \textit{ABC-multi} with the PrimFit solver, UniCo predicts primitives directly and reaches CD 2.18, HD 7.53, and NC 0.935, as reported in \cref{tab:abc_multi}. This corresponds to 40–50\% lower reconstruction error than the strongest competing method, ODGNet with HPNet, while also improving normal consistency. We repeat UniCo’s row across primitive extractor columns to show the gain is consistent. Notably, higher pointwise scores do not guarantee better reconstruction. SymmComplete attains the best pointwise CD and ODGNet the best pointwise F1, yet both still yield lower-quality meshes. As visualized in \cref{fig:completion}, UniCo produces cleaner, solver-aligned primitive layouts.

\begin{figure}[htbp]
 \vspace{-0.5em}
  \centering
  \includegraphics[width=0.95\linewidth]{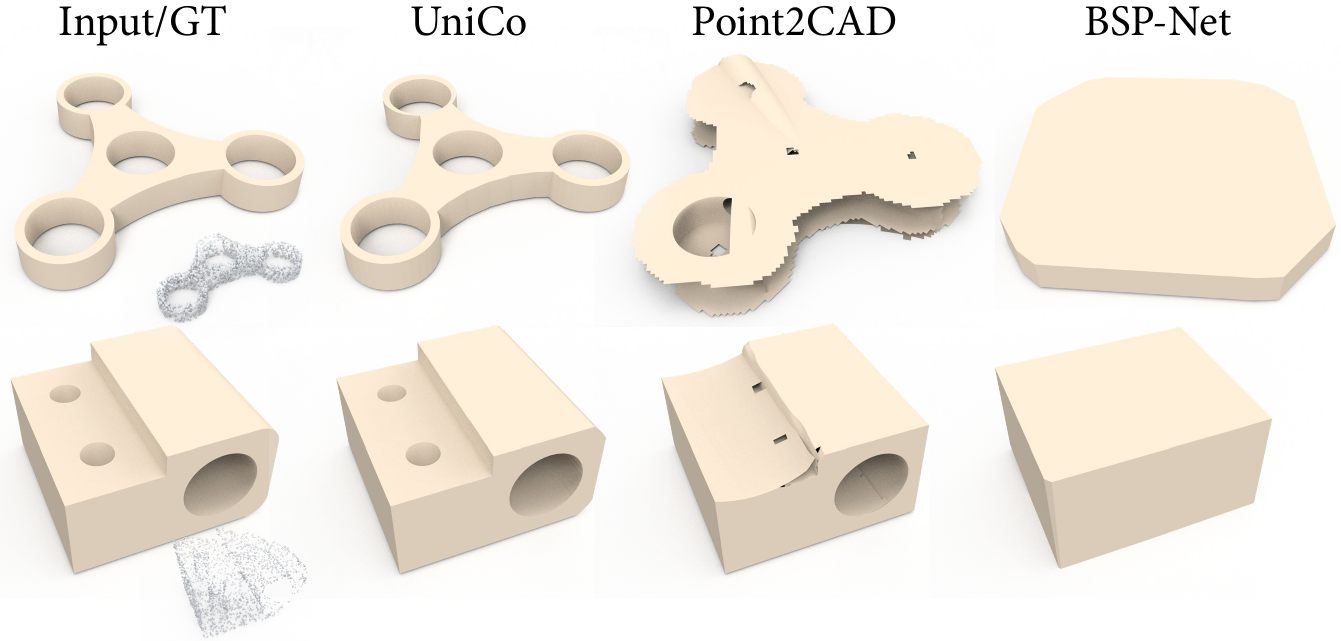}
  \vspace{-0.2em}
  \caption{\textbf{Comparison with reconstruction methods.} Even with completed points with better pointwise metric, both competitors cannot produce detailed and robust reconstructions.}
  \label{fig:reconstruction}
 \vspace{-1em}
\end{figure}

\paragraph{Assembly \vs reconstruction.}
As shown in \cref{tab:reconstruction}, directly reconstructing meshes from partial inputs leads to high geometric errors. Supplying the best pointwise completion (ODGNet) as input to Point2CAD reduces these errors, yet UniCo still achieves substantially better reconstruction, lowering CD by about 36\% and HD by 37\%, while also improving NC. \cref{fig:reconstruction} visualizes the gap: BSP-Net collapses to coarse convex structures, Point2CAD introduces topological breaks, while UniCo’s assembly results preserve detail and maintain topological consistency.

\begin{table}[htbp]
\centering
\vspace{-0.5em}
\caption{\textbf{Comparison with reconstruction methods.} Proc.: R = direct reconstruction; C$\rightarrow$R = completion then reconstruction.}
 \vspace{-0.5em}
\footnotesize
\begin{tabular}{l c cccc}
\toprule
Method & Proc. & {CD $\downarrow$} & {HD $\downarrow$} & {NC $\uparrow$} & {FR $\downarrow$} \\
\midrule
PrimFit \cite{jiang2023primfit} & R & 7.80 & 25.17 & 0.788 & 15.96 \\
Point2CAD \cite{liu2024point2cad} & R & 8.69 & 30.66 & 0.779 & 15.91 \\
BSP-Net \cite{chen2020bspnet} & R & 8.35 & 16.73 & 0.740 & 20.52 \\
CARPRI-Net \cite{yu2022capri} & R & 7.59 &  15.18   & 0.770 & 19.27 \\
\midrule
Point2CAD \cite{liu2024point2cad} & C$\rightarrow$R & \underline{3.39} & \underline{11.96} & \underline{0.833} & \textbf{0.66} \\
\rowcolor{OursRow}
UniCo (ours) & C$\rightarrow$R & \textbf{2.18} & \textbf{7.53} & \textbf{0.935} & \underline{1.49} \\
\bottomrule
\end{tabular}
\label{tab:reconstruction}
 \vspace{-0.5em}
\end{table}

\begin{figure*}[htbp]
  \centering
  \includegraphics[width=0.92\linewidth]{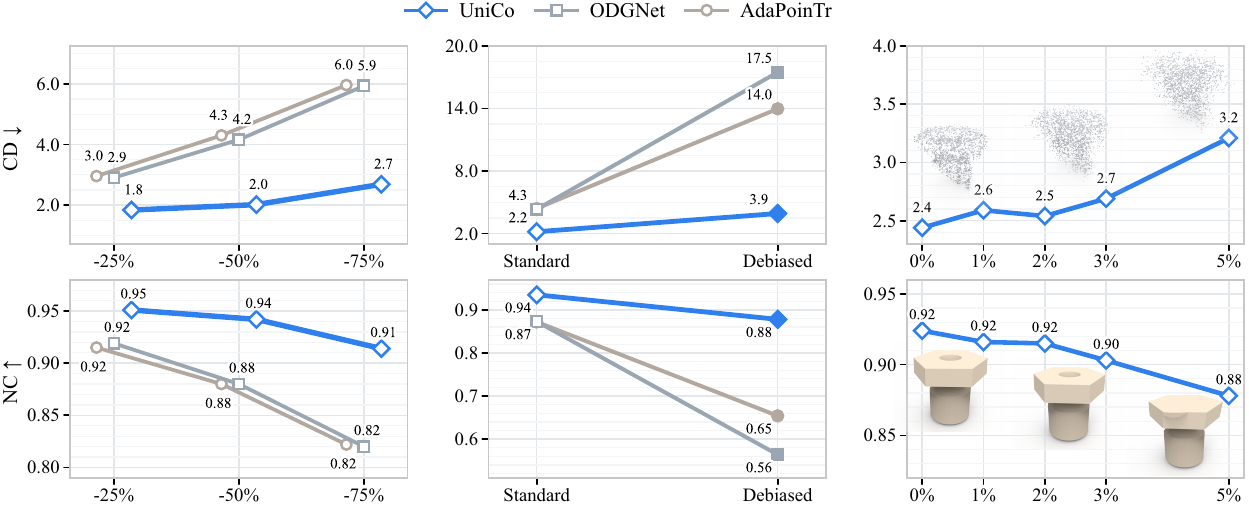}
  \vspace{-0.3em}
  \caption{\textbf{Robustness to missing data, transforms, and noise.} Left: as incompleteness grows from 25\% to 75\%, UniCo maintains lower CD and higher NC than pointwise baselines. Middle: under the debiased normalization protocol~\cite{wang2025simeco}, UniCo remains stable with low pose and scale bias compared to baselines. Right: under Gaussian jitter of 1--3\%, performance degrades gracefully.}
  \label{fig:robustness}
\end{figure*}

\begin{figure}[htbp]
  \centering
  \includegraphics[width=0.88\linewidth]{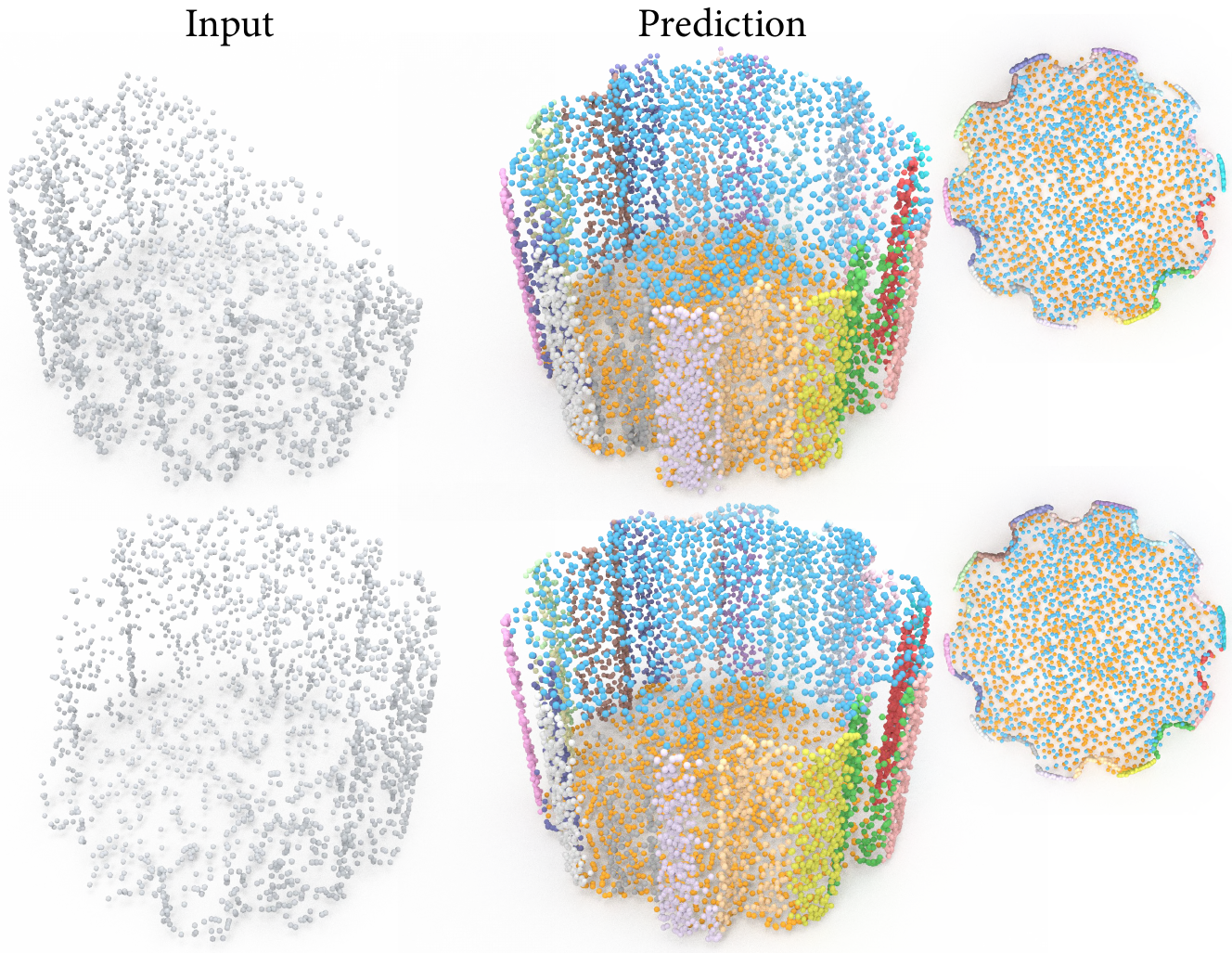}
  \vspace{-0.8em}
  \caption{\textbf{Feature consistency.} From different partial views, UniCo predicts a consistent set of primitives representing the complete geometry. Matching colors denote identical primitive indices.}
  \label{fig:feature}
  \vspace{-1.5em}
\end{figure}

\paragraph{Robustness to missing, transforms, and noise.} We stress-test UniCo under increasing incompleteness, a debiased normalization, and Gaussian jitter, as presented in \cref{fig:robustness}. All networks are trained for 200 epochs. As incompleteness rises from 25\% to 75\%, UniCo’s CD increases only from 1.8 to 2.7, whereas strong pointwise baselines double their error to about 6.0. NC for UniCo drops from 0.95 to 0.91, whereas the baselines drop to 0.82. Under a debiased normalization that removes pose and scale canonicalization~\cite{wang2025simeco}, UniCo degrades to CD 3.9 with NC 0.88, whereas the baselines deteriorate to CD above 14 with NC below 0.65, indicating that UniCo carries substantially less pose and scale bias. With Gaussian noise up to 3\%, UniCo remains stable. At a heavy 5\% noise level, CD and NC degrade to 3.2 and 0.88, respectively. \cref{fig:feature} further shows view-to-view consistency, where different partial inputs yield a consistent primitive set that supports reliable assembly.

\paragraph{Primitive quality analysis.}
To better understand what drives the strong assembly performance, we further evaluate primitive quality following the established protocol~\cite{guo2022complexgen, liu2024point2cad, li2019supervised}. For all competing methods, primitives are extracted with HPNet~\cite{yan2021hpnet} for a feasible and fair comparison. Predicted primitives are then matched to ground truth via Hungarian matching. As shown in \cref{tab:primitive}, we report F-Score@1\% (F1), type accuracy (Type), axis difference (Axis), residual error (Res), and coverage (Cov). Axis measures the plane normal or the symmetry axis of cylinders and cones. Res evaluates the fit on 512 points sampled from the ground truth primitive. Cov is the fraction of these points that lie within a distance of 0.01 from the predicted primitive. UniCo attains the best scores on all primitive metrics, consistent with its stronger reconstruction performance.

\begin{table}[htbp]
\vspace{-0.3em}
\centering
\caption{\textbf{Primitive quality.} Type, Res, and Cov are scaled by 100. Axis is in degrees. UniCo delivers higher-quality primitives.}
\vspace{-0.5em}
\footnotesize
\begin{tabular}{lccccc}
\toprule
Method & F1 $\uparrow$ & Type $\uparrow$  & Axis $\downarrow$ & Res $\downarrow$& Cov $\uparrow$\\
\midrule
GRNet \cite{xie2020grnet}          & 0.215 & 29.65 & 25.97 & 7.12 & 24.41 \\
PoinTr \cite{yu2021pointr}         & 0.509 & 64.41 & 18.24 & 2.86 & 61.60 \\
AdaPoinTr \cite{yu2023adapointr}   & 0.643 & \underline{79.79} & \underline{11.34} & \underline{1.48} & \underline{78.99} \\
ODGNet \cite{cai2024orthogonal}    & \underline{0.659} & 75.52 & 12.24 & 1.78 & 75.85 \\
SymmComplete \cite{yan2025symmcompletion} & 0.629 & 78.48 & 12.72 & 1.54 & 77.60 \\
\rowcolor{OursRow} UniCo (ours)    & \textbf{0.712} & \textbf{94.85} & \textbf{3.29} & \textbf{0.55} & \textbf{92.41} \\
\bottomrule
\end{tabular}
\label{tab:primitive}
\vspace{-1.5em}
\end{table}

\begin{table*}[htbp]
\vspace{-0.3em}
\centering
\caption{\textbf{Comparison on \textit{ABC-plane}.} UniCo achieves leading performance across three assembly solvers for polygonal surface reconstruction.}
\vspace{-0.7em}
\footnotesize
\begin{tabular}{l c cccc|cccc|cccc}
    \toprule
    \multirow{2}{*}{\raisebox{-0.5ex}{Method}} & \multirow{2}{*}{\raisebox{-0.5ex}{Year}}
    & \multicolumn{4}{c}{PolyFit \cite{nan2017polyfit}}   &
      \multicolumn{4}{c}{KSR \cite{bauchet2020kinetic}} & 
      \multicolumn{4}{c}{COMPOD \cite{sulzer2024concise}}\\
    \cmidrule(lr){3-6}\cmidrule(lr){7-10}\cmidrule(lr){11-14}
    & 
    & CD $\downarrow$  & HD $\downarrow$ & NC $\uparrow$ & FR $\downarrow$ 
    & CD $\downarrow$  & HD $\downarrow$  & NC $\uparrow$ & FR $\downarrow$ 
    & CD $\downarrow$  & HD $\downarrow$  & NC $\uparrow$  & FR $\downarrow$\\
    \midrule
    GRNet \cite{xie2020grnet} & 2020
    & 11.98 & 19.84 & 0.769 & 29.61
    & 9.18  & 22.01 & 0.822 & 10.82 
     & 14.19 & 25.72 & 0.738 & 13.15\\

    PoinTr \cite{yu2021pointr} & 2021
    & 10.57 & 16.43 & 0.822 & 25.92
    & 8.14  & 16.33 & 0.780 & 30.90
    & 7.82  & 16.44 & 0.774 & 31.34\\

    AdaPoinTr \cite{yu2023adapointr} & 2023
    & 3.16 & 7.36 & 0.920 & 5.89
    & 3.24 & 8.86 & 0.927 & 0.27 
    & 3.25 & 8.84 & 0.921 & 1.32\\

    ODGNet \cite{cai2024orthogonal} & 2024
    & 2.73 & 6.41 & 0.933 & 4.28
    & 2.90 & 8.56 & 0.934 & 0.36 
    & 3.22 & 8.01 & 0.927 & 1.05\\

    SymmComplete \cite{yan2025symmcompletion} & 2025
    & 3.21 & 9.15 & 0.920 & 4.27
    & 4.23 & 14.21 & 0.907 & 0.38 
     & 4.56 & 14.02 & 0.895 & 4.14\\

    PaCo \cite{chen2025parametric} & 2025
    & \underline{1.87} & \textbf{4.09} & \underline{0.943} & \textbf{0.48}
    & \underline{1.91} & \textbf{4.14} & \underline{0.940} & \underline{0.25} 
    & \underline{1.94} & \underline{4.42} & \underline{0.940} & \underline{0.25}\\

    \rowcolor{OursRow}
    UniCo (ours) & -
    & \textbf{1.69} & \underline{4.28} & \textbf{0.953} & \underline{0.69}
    & \textbf{1.78} & \underline{4.59} & \textbf{0.951} & \textbf{0.00} 
    & \textbf{1.63} & \textbf{4.27} & \textbf{0.952} & \textbf{0.00}\\
    \bottomrule
\end{tabular}
\label{tab:abc_plane}
\vspace{-1.7em}
\end{table*}


\subsection{Planar Primitive Results}
\vspace{-0.8em}
\paragraph{Specialization with minimal change.}
We reduce the semantic head in \cref{eq:type-softmax} to a binary classifier and replace the geometry head that predicts quadric coefficients with one that predicts plane parameters, leaving the rest of the architecture and training unchanged, which results in a strong model for plane-dominant scenarios. On \textit{ABC-plane}, UniCo achieves the lowest CD and highest NC across PolyFit, COMPOD, and KSR, as presented in \cref{tab:abc_plane}. HD is best with COMPOD and second best with PolyFit and KSR, and the failure rate is even zero with both COMPOD and KSR. The planar primitives facilitate polygonal surface reconstruction more effectively than unstructured, pointwise completions, so the reconstructed surfaces are clean and well aligned.

\begin{figure*}[htbp]
  \centering\includegraphics[width=0.88\linewidth]{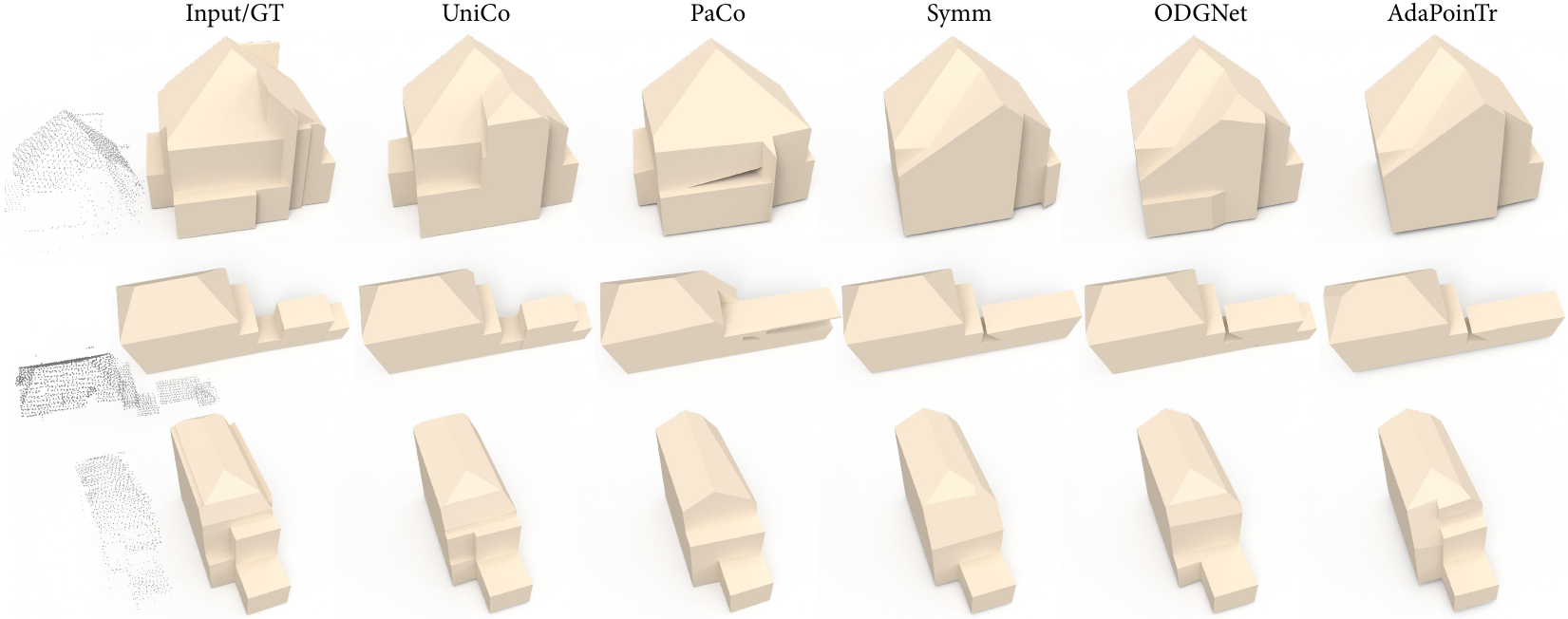}
   \vspace{-0.2em}
  \caption{\textbf{Building reconstruction from real LiDAR scans.} UniCo recovers superstructures with higher fidelity.}
  \label{fig:building}
  \vspace{-0.9em}
\end{figure*}

\paragraph{Performance on real scans.}
On \textit{Building-PCC}, UniCo delivers strong reconstruction results, as presented in \cref{tab:building}. It achieves the best CD across all three solvers and the lowest failure rates. With PolyFit it also reaches the highest NC and a competitive HD, and with COMPOD it leads all three metrics. \cref{fig:building} presents qualitative results with PolyFit, showing roof superstructures with fewer distortions and preserving structural integrity across diverse architectural styles.

\begin{table*}[htbp]
\centering
\caption{\textbf{Comparison on \textit{Building-PCC}.} On real LiDAR scans, UniCo produces the most reliable reconstructions across three solvers.}
\vspace{-0.7em}
\footnotesize
\begin{tabular}{lcccc|cccc|cccc}
    \toprule
    \multirow{2}{*}{\raisebox{-0.5ex}{Method}} 
    & \multicolumn{4}{c}{PolyFit \cite{nan2017polyfit}}   &
      \multicolumn{4}{c}{KSR \cite{bauchet2020kinetic}}  & 
      \multicolumn{4}{c}{COMPOD \cite{sulzer2024concise}}\\
    \cmidrule(lr){2-5}\cmidrule(lr){6-9}\cmidrule(lr){10-13}
    & CD $\downarrow$  & HD $\downarrow$ & NC $\uparrow$ & FR $\downarrow$ 
    & CD $\downarrow$  & HD $\downarrow$  & NC $\uparrow$ & FR $\downarrow$ 
    & CD $\downarrow$  & HD $\downarrow$  & NC $\uparrow$  & FR $\downarrow$\\
    \midrule
    AdaPoinTr \cite{yu2023adapointr}
    & 4.87 & 10.61 & 0.934 & 0.85
    & 4.50 & \textbf{9.56} & 0.939 & \underline{0.07}
    & 5.92 & 18.94 & 0.917 & \underline{0.15}\\

    ODGNet \cite{cai2024orthogonal}
    & \underline{3.97} & \textbf{9.09} & \underline{0.947} & 0.87
    & \underline{4.41} & \underline{9.88} & \textbf{0.947} & 0.61 
     & \underline{4.71} & 13.81 & \underline{0.940} & 1.77\\

    SymmComplete \cite{yan2025symmcompletion}
    & 7.55 & 14.74 & 0.893 & 24.98
    & 13.09 & 24.45 & 0.803 & 38.94 
     & 9.73 & 22.54 & 0.878 & 20.48\\

    PaCo \cite{chen2025parametric}
    & 4.89 & 10.74 & 0.932 & \underline{0.54}
    & 4.47 & 10.20 & 0.934 & 0.17 
     & \underline{4.71} & \underline{11.73} & 0.932 & \textbf{0.00}\\

    \rowcolor{OursRow}
    UniCo (ours)
    & \textbf{3.84} & \underline{9.18} & \textbf{0.949} & \textbf{0.39}
    & \textbf{4.19} & 10.67 & \underline{0.944} & \textbf{0.00} 
    & \textbf{4.08} & \textbf{10.81} & \textbf{0.941} & \textbf{0.00}\\
    \bottomrule
\end{tabular}
\label{tab:building}
\vspace{-0.5em}
\end{table*}

\paragraph{Joint \vs cascaded.}
Compared with the cascaded baseline PaCo, UniCo yields cleaner primitives with better support and hence more reliable reconstructions. On \textit{ABC-plane}, it improves CD and NC across all three solvers, \eg with COMPOD CD drops from 1.94 to 1.63 and FR from 0.25 to 0.00, as reported in \cref{tab:abc_plane}. \textit{Building-PCC} is more challenging, with many small and detailed structures, yet UniCo again excels across all metrics, \eg with PolyFit the CD decreases from 4.89 to 3.84, as reported in \cref{tab:building}. These results highlight that the coordinated pathways for learning primitives and points jointly are more effective than a cascaded design. Additional analysis is provided in the \underline{Appendix}.

\begin{table}[ht]
\vspace{-0.5em}
\centering
\caption{\textbf{Ablation results.} ``$\dagger$'' parameter head is required by solvers that depend on explicit parameters.}
\vspace{-0.7em}
\footnotesize
\setlength{\tabcolsep}{6pt}
\renewcommand{\arraystretch}{1.1}
\begin{tabular}{l rr}
\toprule
Variant & {CD $\downarrow$} & {NC $\uparrow$} \\
\midrule
\multicolumn{3}{@{}l}{\emph{Heads}} \\
\quad no param. head$^\dagger$   & 2.52 {\scriptsize($-0.08$)} & 0.921 {\scriptsize($-0.003$)} \\
\quad no prim. chamfer           & 2.53 {\scriptsize($-0.09$)} & 0.920 {\scriptsize($-0.004$)} \\

\addlinespace[2pt]
\multicolumn{3}{@{}l}{\emph{Membership}} \\
\quad CE-only memb.        & 2.53 {\scriptsize($-0.09$)} & 0.923 {\scriptsize($-0.001$)} \\
\quad dice-only memb.       & 2.66 {\scriptsize($-0.22$)} & 0.914 {\scriptsize($-0.010$)} \\
\addlinespace[2pt]
\multicolumn{3}{@{}l}{\emph{Training}} \\
\quad no online target      & 12.22 {\scriptsize($-9.78$)} & 0.631 {\scriptsize($-0.293$)} \\
\quad two-stage training    & 2.55 {\scriptsize($-0.11$)} & 0.919 {\scriptsize($-0.005$)} \\
\rowcolor{OursRow}
UniCo (ours) & \textbf{2.44}\phantom{{\scriptsize($00000$)} } &
               \textbf{0.924}\phantom{ {\scriptsize($-0000$)} } \\
\bottomrule
\end{tabular}
\label{tab:ablation}
\vspace{-1em}
\end{table}

\subsection{Ablations}
We ablate core modules in heads, membership, and training, as summarized in \cref{tab:ablation}, and train all variants for 200 epochs. Dropping the parameter head changes scores only slightly, but it remains necessary for assembly solvers that require explicit parameters (\eg, normals). Removing primitive-wise Chamfer slightly increases CD and lowers NC. For membership supervision, combining cross-entropy and Dice works best, while using only one increases CD and reduces NC marginally. Training designs are most critical. Removing online target supervision is catastrophic, pushing CD to about five times that of the full model. Switching to a two-stage pipeline, where we first train the point pathway and then the primitive pathway, also increases CD and reduces NC. 

\vspace{-0.4em}
\section{Conclusion and Discussion}
\label{sec:conclusion}

We rethought how primitives and points should interact for structured shape completion and introduced UniCo, a unified model that predicts assembly-ready primitives through primitive proxies. On mixed-type, plane-only, and real airborne LiDAR benchmarks, UniCo achieves state-of-the-art performance, establishing an effective recipe for structured 3D understanding from incomplete data.

\paragraph{Limitations.}
UniCo is tailored for shape completion aimed at primitive assembly and therefore prioritizes assembly-ready structure over pointwise fidelity. It is not intended for highly unstructured geometry where primitive abstraction provides limited benefit, and its final reconstruction quality depends on the downstream solver. Within this scope, however, UniCo reliably learns structures that assemble correctly.

\paragraph{Future work.}
Our method extends to richer primitive families. We observe that primitive proxies develop consistent proxy-level semantics, with specific proxies representing the same object parts even without explicit supervision. Future directions include exploiting these emergent correspondences for part-aware assembly and scaling UniCo to larger scenes.

\section*{Acknowledgments}
This work was supported by TUM Georg Nemetschek Institute under the AI4TWINNING project. We thank Jing-En Jiang for support with PrimFit and Liangliang Nan for helpful discussions.

\section*{\LARGE Appendix}
In the appendix, we provide instructions for reproducing our results (\cref{app:reproducibility}), detailed implementation settings (\cref{app:implementation}), extended experimental analyses (\cref{app:analyses}), and further details on the datasets (\cref{app:datasets}) and metrics (\cref{app:metrics}).

\renewcommand{\thefigure}{S\arabic{figure}}
\setcounter{figure}{0} 
\renewcommand{\thetable}{S\arabic{table}}
\setcounter{table}{0} 
\renewcommand{\theequation}{S\arabic{equation}}
\setcounter{equation}{0} 

\renewcommand\thesection{\Alph{section}}
\setcounter{section}{0}


\section{Reproducibility}
\label{app:reproducibility}

The code repository and demo are publicly accessible via the project page\footnote{\url{https://unico-completion.github.io}}. Detailed instructions for setup and running the code are described in the repository’s \texttt{README.md} file.

\section{Implementation Details}
\label{app:implementation}

The point pathway follows the AdaPoinTr backbone~\cite{yu2023adapointr} with its default depth and hyperparameters. Input points are grouped into local neighborhoods, encoded with self-attention, and decoded from learned point queries. The decoder produces \(U = 512\) shape features, which a lightweight reconstruction head expands into 8{,}192 completed points. As in AdaPoinTr, we use denoising queries during training for an auxiliary denoising loss and drop them at inference.
The primitive pathway consumes the same \(U\) shape features. We use \(K = 40\) learnable primitive queries, contextualized by a 4-layer Transformer decoder with 8 attention heads and a hidden size of 128. For each query, a prediction head outputs a primitive type, a soft mask over the 512 shape features, and 10 coefficients of a homogeneous quadric.

UniCo is implemented in PyTorch and optimized using the AdamW optimizer with an initial learning rate of $2 \times 10^{-3}$, a weight decay of $5 \times 10^{-4}$, and a learning rate decay of 0.9 every 20 epochs. During inference, primitives with confidence \( s_k > 0.5\) are retained. For the loss terms in \cref{eq:matching}, we empirically set  \( \alpha_2  = 0.125 \),  \( \alpha_3 = 1 \) and  \( \lambda  =  0.05 \). To mitigate class imbalance, we set
\begin{equation}
\alpha_1 =
\begin{cases}
0.05, & \text{if } c_g \neq \emptyset, \\
0.01, & \text{otherwise.}
\end{cases}
\end{equation}
where $c_g$ denotes the ground-truth primitive type. On \textit{ABC-multi}, shapes contain on average about 8 primitives while we use $K = 40$ proxies, so the ratio for no-object \vs valid types roughly matches the expected proportion and keeps their aggregate loss contributions comparable. The impact of key hyperparameters is analyzed in \cref{app:analyses}.



\section{Additional Analyses}
\label{app:analyses}

\subsection{More Ablations}

\paragraph{Number of primitive proxies.}
We vary the number of primitive proxies \(K \in \{30, 40, 50\}\) and the no-object weight $\alpha_1(c_g = \emptyset)$ while keeping all other settings fixed. As reported in \cref{tab:nproxy}, \(K = 40\) with a no-object weight of 0.01 achieves the best results.

\begin{table}[htb]
    \centering
    \caption{\textbf{Effect of proxy count and no-object weight.} Parameter counts (in millions) only include the primitive pathway \(f_\mathrm{primitive}\).}
    \footnotesize
    \begin{tabular}{ccccccc}
        \toprule
        \(K\) & $\alpha_1 (c_g = \emptyset)$ & Params (M) & CD $\downarrow$ & HD $\downarrow$ & NC $\uparrow$ & FR $\downarrow$ \\
        \midrule
        30 & 0.01 & 0.901 & 2.47 & 8.70 & 0.923 & 1.98 \\
        \rowcolor{OursRow}
        40 & 0.01 & 0.902 & 2.44 & 8.80 & 0.924 & 1.83 \\
        40 & 0.05 & 0.902 & 2.73 & 9.60 & 0.915 & 1.96 \\
        50 & 0.01 & 0.904 & 2.48 & 8.94 & 0.920 & 1.59 \\
        \bottomrule
    \end{tabular}
    \label{tab:nproxy}
\end{table}

\paragraph{Confidence threshold.}
We assess the effect of the confidence threshold applied to the scores \(s_k\) in \cref{eq:final-score}, varying the pruning value in \(\{0.3, 0.5, 0.7\}\) at inference time. As summarized in \cref{tab:confidence}, performance is fairly stable across thresholds, and \(s_k > 0.5\) gives the best overall results.

\begin{table}[htb]
    \centering
    \caption{\textbf{Effect of confidence threshold.} 
    Changing the pruning cutoff on $s_k$ within a reasonable range has little effect on performance, while using $s_k > 0.5$ gives the best results.}
    \footnotesize
    \begin{tabular}{ccccc}
        \toprule
        $s_k > $ & CD $\downarrow$ & HD $\downarrow$ & NC $\uparrow$ & FR $\downarrow$ \\
        \midrule
        0.3 & 2.48 & 8.81  & 0.923 &  1.89\\
        \rowcolor{OursRow}
        0.5 & 2.44 & 8.80 & 0.924 & 1.83 \\
        0.7 & 2.48  &8.84  &0.923  &  1.80\\
        \bottomrule
    \end{tabular}
    \label{tab:confidence}
\end{table}



\paragraph{Analytic \vs fitted primitives.}
In \cref{tab:analysis_vs_fitting}, we compare primitives obtained directly from analytic quadric parameters with primitives obtained by fitting quadrics to the completed points. Since the assignments are identical, F1 and type accuracy remain unchanged. Analytic parameters achieve a lower axis error, whereas fitted primitives reduce the residual error and slightly improve coverage. For consistency with competing methods that rely on fitted primitives, the main paper reports the fitted variant.

\begin{table}[htbp]
\centering
\footnotesize
\begin{tabular}{l >{\color{gray}}c >{\color{gray}}c cccc}
\toprule
Source         & F1 $\uparrow$ & Type $\uparrow$ & Axis $\downarrow$ & Res $\downarrow$ & Cov $\uparrow$ \\
\midrule
Analytic       & 0.712 & 94.85 & {2.71} & 0.70 & 92.16 \\
\rowcolor{OursRow}
Fitted         & 0.712 & 94.85 & 3.29         & {0.55} & {92.41} \\
\bottomrule
\end{tabular}
\caption{\textbf{Primitive quality for analytic \vs fitted sources.} Fitting quadrics to completed points improves residual error and coverage at a cost in axis error.}
\label{tab:analysis_vs_fitting}
\end{table}

\begin{figure}[htbp]
    \centering\includegraphics[width=0.95\linewidth]{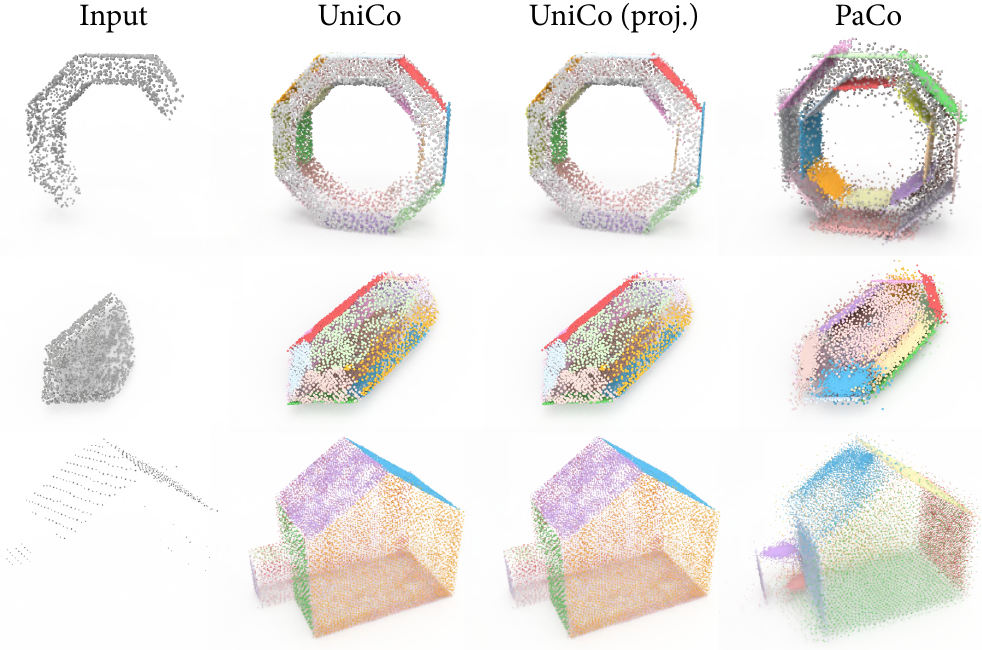}
    \caption{Qualitative primitive comparison on \textit{ABC-plane} and \textit{Building-PCC}. UniCo recovers more uniform point distributions and cleaner, well-aligned primitive structures.}
    \label{fig:paco}
\end{figure}

\paragraph{Projection.}
\cref{tab:projection} compares UniCo with and without a projection-based post-processing step that projects completed points onto their predicted primitives to enforce stricter planar geometry, with PaCo~\cite{chen2025parametric} as a reference. On both \textit{ABC-plane} and \textit{Building-PCC}, UniCo without projection already improves over PaCo, and projection brings only modest additional gains. For fairness, the main paper therefore reports UniCo without projection and treats the projected variant as an optional refinement for slightly sharper surfaces. \cref{fig:paco} shows that PaCo tends to under-represent small primitives and fine details, whereas UniCo produces more uniform point distributions and cleaner, well-aligned primitive layouts across scales. Projection mainly sharpens surfaces and does not change this overall qualitative picture.

\begin{table}[htb]
\centering
\caption{\textbf{Effect of projection-based refinement.} UniCo already outperforms PaCo on \textit{ABC-plane} and \textit{Building-PCC}, and projection yields only marginal gains.}
\footnotesize
\setlength{\tabcolsep}{5pt}
\begin{subtable}{\linewidth}
\centering
\caption{\textit{ABC-plane}}
\begin{tabular}{lcccc}
\toprule
Method & CD $\downarrow$ & HD $\downarrow$ & NC $\uparrow$ & FR $\downarrow$ \\
\midrule
PaCo \cite{chen2025parametric}   & {1.87} & {4.09} & {0.943} & {0.48} \\
\rowcolor{OursRow}
UniCo                            & {1.69} & {4.28} & {0.953} & {0.69} \\
UniCo (proj.)                & 1.67 & 4.29 & 0.955 & 0.55 \\
\bottomrule
\end{tabular}
\vspace{1.0em}
\end{subtable}

\begin{subtable}{\linewidth}
\centering
\caption{\textit{Building-PCC}}
\begin{tabular}{lcccc}
\toprule
Method & CD $\downarrow$ & HD $\downarrow$ & NC $\uparrow$ & FR $\downarrow$ \\
\midrule
PaCo \cite{chen2025parametric}   & 4.89 & 10.74 & 0.932 & {0.54} \\
\rowcolor{OursRow}
UniCo                            & {3.84} & {9.18} & {0.949} & {0.39} \\
UniCo (proj.)                & 3.83 & 9.06 & 0.949 & 0.17 \\
\bottomrule
\end{tabular}
\end{subtable}
\label{tab:projection}
\end{table}

\paragraph{Robustness across backbones.}
We additionally swap the point completion backbone $f_{\mathrm{point}}$ to PoinTr \cite{yu2021pointr} or ODGNet \cite{cai2024orthogonal}. UniCo retains \textit{consistent} gains over the corresponding completion+extractor baselines. This shows that the improvements are driven by UniCo's primitive pathway and persist across backbones.

\begin{table}[ht]
    \centering
    \caption{\textbf{Robustness across backbones.} UniCo retains consistent gains over the corresponding completion+extractor baselines.}
    \footnotesize
    \setlength{\tabcolsep}{3.6pt}
    \begin{tabular}{l|ccc|ccc}
        \toprule
        \multirow{2}{*}{Method} &
        \multicolumn{3}{c|}{$f_{\mathrm{point}}=\text{PoinTr}$ \cite{yu2021pointr}} &
        \multicolumn{3}{c}{$f_{\mathrm{point}}=\text{ODGNet}$} \cite{cai2024orthogonal}\\
        & CD$\downarrow$ & NC$\uparrow$ & FR$\downarrow$
        & CD$\downarrow$ & NC$\uparrow$ & FR$\downarrow$ \\
        \midrule
        Compl.+Extr. & 6.58 & 0.791 & 11.27 & 4.33 & 0.873 & 7.41 \\
        \rowcolor{OursRow}
        UniCo (ours) & 2.64 & 0.915 &  2.49 & 2.47 & 0.922 & 0.31 \\
        \bottomrule
    \end{tabular}
    \label{tab:backbone}
\end{table}

\vspace{-0.5em}
\subsection{Transferability}

We train UniCo separately on the \textit{ABC-multi} and \textit{ABC-plane} and evaluate both on the plane-only split. To avoid data leakage, we exclude test shapes that also appear in the mixed-type training set. As shown in \cref{tab:transferability}, the model trained on the mixed-type split exhibits only moderate degradation across all metrics and still achieves strong performance, indicating that UniCo transfers well from mixed-type to plane-only data.

\begin{table}[htbp]
\centering
\vspace{-0.5em}
\caption{\textbf{Transferability.} UniCo trained on different datasets and evaluated on \textit{ABC-plane}, excluding overlapping samples.}
\footnotesize
\setlength{\tabcolsep}{6pt}
\renewcommand{\arraystretch}{1.1}
\begin{tabular}{@{}lcccc@{}}
\toprule
\multicolumn{1}{c}{Training $\rightarrow$ Evaluation} &
CD $\downarrow$ & HD $\downarrow$ & NC $\uparrow$ & FR $\downarrow$ \\
\midrule
ABC-plane $\rightarrow$ ABC-plane & 1.70 & 4.35 & 0.953 & 1.16 \\
ABC-multi $\rightarrow$ ABC-plane & 2.06 & 5.36 & 0.942 & 2.13 \\
\bottomrule
\end{tabular}
\label{tab:transferability}
\end{table}

\begin{figure}[ht]
    \centering
    \includegraphics[width=0.99\linewidth]{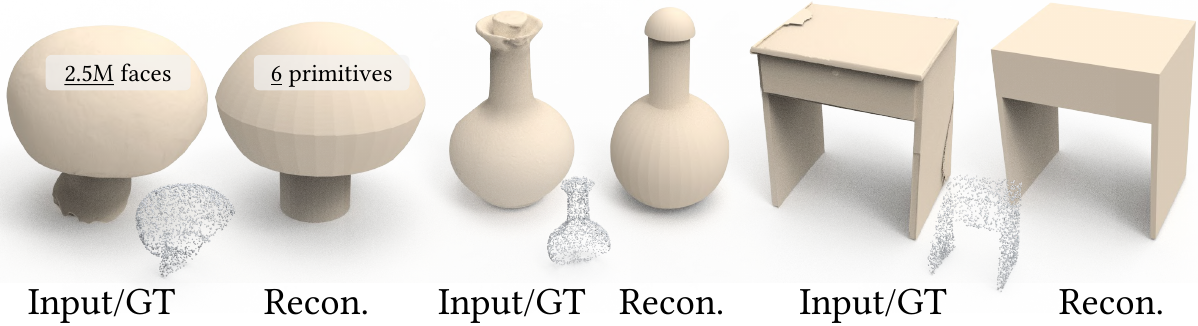}
    \caption{\textbf{OOD generalization via approximation.} UniCo trained on synthetic data with a fixed primitive set generalizes to real-world point clouds with unseen geometry by approximation.}
    \label{fig:approximation}
\end{figure}

Moreover, UniCo \textit{generalizes} to unseen freeform geometries in \textit{real} OmniObject3D scans \cite{wu2023omniobject3d} \textit{without} training on them, by compactly approximating them with the available primitives, as shown in \cref{fig:approximation}.

\vspace{1em}


\subsection{Computational Efficiency}

\cref{tab:efficiency} reports per-scan runtimes on \textit{ABC-plane}. UniCo processes each partial input in 27.6\,ms end-to-end, roughly twice as fast as the strongest pointwise competitor, ODGNet~\cite{cai2024orthogonal}, and faster than the structured competitor PaCo~\cite{chen2025parametric}. All other completion methods, except PaCo and UniCo, require an additional primitive extraction stage. To quantify this overhead, we evaluate RANSAC~\cite{schnabel2007efficient}, HPNet~\cite{yan2021hpnet}, and PTv3~\cite{wu2024ptv3} on 100 randomly selected ABC-multi samples completed by ODGNet. UniCo is the fastest end-to-end pipeline.

\begin{table*}[htbp]    
\centering
\caption{\textbf{Runtime and complexity.} All completion methods are measured on a single A40 GPU, excluding the first iteration to ensure steady-state measurements. For methods requiring primitive extraction, we additionally report RANSAC, HPNet, and PTv3 post-processing costs, evaluated on 100 random \textit{ABC-multi} samples. ``Total Params'' and ``Total Latency'' refer to the parameter count and latency of the full pipeline. Params are reported in millions and all times are in milliseconds.}
\footnotesize
\setlength{\tabcolsep}{5pt}

\begin{tabular}{
    lGG
    cc!{\vrule width .6pt}
    cc!{\vrule width .6pt}
    cc
}
\toprule
\multirow{2}{*}{Method} &
\multirow{2}{*}{Params} &
\multirow{2}{*}{Latency} &
\multicolumn{2}{c}{RANSAC \cite{schnabel2007efficient}} &
\multicolumn{2}{c}{HPNet \cite{yan2021hpnet}} &
\multicolumn{2}{c}{PTv3 \cite{wu2024ptv3}} \\
\cmidrule(lr){4-5}
\cmidrule(lr){6-7}
\cmidrule(lr){8-9}
& & 
& Total Params & Total Latency
& Total Params & Total Latency
& Total Params & Total Latency
\\
\midrule
AdaPoinTr \cite{yu2023adapointr}
  & 32.5 & \underline{23.9}
  & 32.5 & 124.9
  & 33.8 & 1280.9
  & 208.5 & 169.3 \\
ODGNet \cite{cai2024orthogonal}
  & \textbf{11.5} & 53.6
  & \textbf{11.5} & 154.6
  & \textbf{12.8} & 1310.6
  & 187.5 & 199.0 \\
SymmComplete \cite{yan2025symmcompletion}
  & \underline{13.3} & \textbf{20.0}
  & \underline{13.3} & 121.0
  & \underline{14.6} & 1277.0
  & 189.4 & 165.4 \\
PaCo \cite{chen2025parametric}
  & 41.4 & 29.8
  & 41.4 & \underline{29.8}
  & 41.4 & \underline{29.8}
  & \underline{41.4} & \underline{29.8} \\
\rowcolor{OursRow}
UniCo (ours)
  & 33.4 & 27.6
  & 33.4 & \textbf{27.6}
  & 33.4 & \textbf{27.6}
  & \textbf{33.4} & \textbf{27.6} \\
\bottomrule
\end{tabular}
\label{tab:efficiency}
\end{table*}
\vspace{0.5em}

\subsection{Failure cases}
\begin{figure}[ht]
    \centering
    \includegraphics[width=1.0\linewidth]{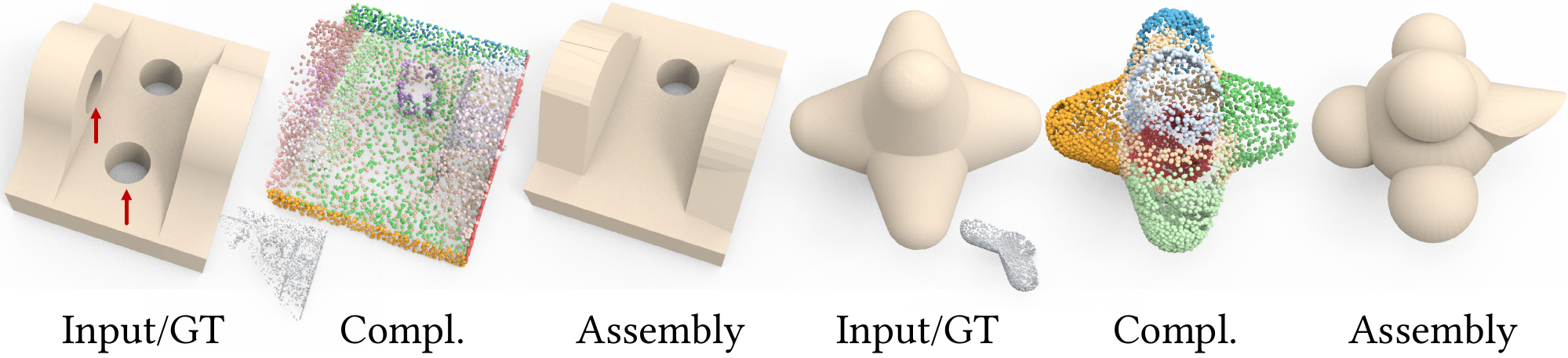}
    \caption{\textbf{Failure cases.} Typical failure cases involve missing small primitives and misclassified semantics.}
    \label{fig:failure}
    \vspace{0.5em}
\end{figure}

Typical failure modes involve missing small primitives and misclassified semantics, with examples shown in \cref{fig:failure}. These cases are overall rare and occur less often than in competing methods (see \cref{tab:primitive}).

\section{Datasets}
\label{app:datasets}

For \textit{ABC-multi}, we randomly select 30{,}000 single-piece watertight CAD models from the ABC dataset~\cite{koch2019abc}, covering a broad spectrum of primitive configurations, from simple shapes with only a few primitives to complex assemblies with several tens of parts. Planes and cylinders dominate the primitive inventory, while cones and spheres provide additional geometric variety. This yields a diverse, structured benchmark for primitive-based completion. A quantitative summary is provided in \cref{tab:dataset_stats}.

\begin{table}[ht]
\caption{\textbf{Primitive statistics on \textit{ABC-multi}.} Primitive counts, per-sample statistics, and type compositions for 30{,}000 CAD models.}
\centering
\footnotesize
\setlength{\tabcolsep}{6pt}
\begin{tabular}{l r}
\toprule
\textbf{Statistic} & \textbf{Value} \\
\midrule
Samples & 30{,}000 \\
Primitive instances & 219{,}477 \\
\addlinespace[0.3em]
\multicolumn{2}{l}{\textit{Primitive instances by type}} \\
\quad Plane & 156{,}558 (71.3\%) \\
\quad Cylinder & 55{,}349 (25.2\%) \\
\quad Cone & 6{,}541 (3.0\%) \\
\quad Sphere & 1{,}029 (0.5\%) \\
\addlinespace[0.3em]
\multicolumn{2}{l}{\textit{Primitives per sample}} \\
\quad Min / median / max & 2 / 7 / 38 \\
\quad 25th / 75th percentile & 4 / 10 \\
\quad Mean & 7.32 \\
\addlinespace[0.3em]
\multicolumn{2}{l}{\textit{Type composition}} \\
\quad Plane & 5{,}800 (19.3\%) \\
\quad Cylinder & 19 (0.1\%) \\
\quad Cone & 87 (0.3\%) \\
\quad Sphere & 15 (0.1\%) \\
\quad Plane+Cylinder & 20{,}404 (68.0\%) \\
\quad Plane+Cone & 448 (1.5\%) \\
\quad Plane+Sphere & 139 (0.5\%) \\
\quad Cone+Cylinder & 22 (0.1\%) \\
\quad Cylinder+Sphere & 29 (0.1\%) \\
\quad Plane+Cone+Cylinder & 2{,}538 (8.5\%) \\
\quad Plane+Cylinder+Sphere & 339 (1.1\%) \\
\quad Plane+Cone+Cylinder+Sphere & 121 (0.4\%) \\
\quad Other mixed combinations & 39 (0.1\%) \\
\bottomrule
\end{tabular}
\label{tab:dataset_stats}
\vspace{-1.0em}
\end{table}

In addition, we use \textit{ABC-plane}~\cite{chen2025parametric}, a plane-only subset of ABC, and \textit{Building-PCC}~\cite{gao2024building}, an airborne LiDAR dataset of roughly 50{,}000 urban buildings with noise and occlusions. These datasets complement \textit{ABC-multi} by providing plane-only CAD assemblies and real-world scans for evaluation.

\section{Metrics}
\label{app:metrics}

For failed reconstructions, we follow the established protocol~\cite{chen2025parametric} and evaluate against the unit cube. Below we detail the additional primitive-level metrics used in \cref{tab:primitive}, following established practice~\cite{guo2022complexgen, liu2024point2cad, li2019supervised}. We denote a matched primitive pair by \((k^{*}, g^{*}) \in \mathcal{M}\):

{\footnotesize \begin{equation*} \begin{aligned} \textbf{F1:}\quad &\frac{1}{|\mathcal{M}|}\sum_{(k^*,g^*)\in\mathcal{M}} \mathrm{F}_1(k^*, g^*), \\[-0.2em] \textbf{Type:}\quad &\frac{1}{|\mathcal{M}|}\sum_{(k^*,g^*)\in\mathcal{M}} \mathbbm{1}\{\hat{c}_{k^*}=c_{g^*}\}, \\[-0.2em] \textbf{Axis:}\quad &\frac{\sum_{(k^*,g^*)\in\mathcal{M}}\mathbbm{1}\{\hat{c}_{k^*}=c_{g^*}\}\, \arccos\!\left\langle n_{k^*},\, n_{g^*}\right\rangle} {\sum_{(k^*,g^*)\in\mathcal{M}}\mathbbm{1}\{\hat{c}_{k^*}=c_{g^*}\}}, \\[-0.2em] \textbf{Res:}\quad &\frac{\sum_{(k^*,g^*)\in\mathcal{M}}\mathbbm{1}\{\hat{c}_{k^*}\neq \emptyset\}\, \mathbb{E}_{\mathbf{x}\sim U(\theta_{g^*})} \operatorname{D}(\mathbf{x},\theta_{k^*})} {\sum_{(k^*,g^*)\in\mathcal{M}}\mathbbm{1}\{\hat{c}_{k^*}\neq \emptyset\}}, \\[-0.2em] \textbf{Cov:}\quad &\frac{\sum_{(k^*,g^*)\in\mathcal{M}}\mathbbm{1}\{\hat{c}_{k^*}\neq \emptyset\}\, \mathbb{E}_{\mathbf{x}\sim U(\theta_{g^*})} \mathbbm{1}\{\operatorname{D}(\mathbf{x},\theta_{k^*}) < \epsilon\}}{\sum_{(k^*,g^*)\in\mathcal{M}}\mathbbm{1}\{\hat{c}_{k^*}\neq \emptyset\}}, \end{aligned} \end{equation*}}\noindent
where \(\operatorname{D}(\mathbf{x},\theta_{k^*})\) denotes the point-to-primitive distance, and \(\mathbf{x}\sim U(\theta_{g^*})\) indicates uniform sampling over the bounded surface of the ground-truth primitive.


{
    \small
    \bibliographystyle{ieeenat_fullname}
    \bibliography{main}
}

\end{document}